\documentclass[runningheads]{llncs}

\usepackage{graphicx}
\usepackage{array} 
\usepackage{multirow} 
\usepackage{url}
\usepackage{ctable}
\usepackage{cite}
\usepackage{subfig}
\usepackage{appendix}

\begin{document}

\title{Assessing the Generalizability of Deep Neural Networks-Based Models for Black Skin Lesions}

\author{Luana Barros \and Levy Chaves \and Sandra Avila}
\authorrunning{L. Barros et al.}
\institute{Recod.ai Lab, Institute of Computing, University of Campinas, Brazil\\
\email{sandra@ic.unicamp.br}}

\maketitle             
\begin{abstract}
Melanoma is the most severe type of skin cancer due to its ability to cause metastasis. It is more common in black people, often affecting acral regions: palms, soles, and nails. Deep neural networks have shown tremendous potential for improving clinical care and skin cancer diagnosis. Nevertheless, prevailing studies predominantly rely on datasets of white skin tones, neglecting to report diagnostic outcomes for diverse patient skin tones. In this work, we evaluate supervised and self-supervised models in skin lesion images extracted from acral regions commonly observed in black individuals. Also, we carefully curate a dataset containing skin lesions in acral regions and assess the datasets concerning the Fitzpatrick scale to verify performance on black skin. Our results expose the poor generalizability of these models, revealing their favorable performance for lesions on white skin. Neglecting to create diverse datasets, which necessitates the development of specialized models, is unacceptable. Deep neural networks have great potential to improve diagnosis, particularly for populations with limited access to dermatology. However, including black skin lesions is necessary to ensure these populations can access the benefits of inclusive~technology.

\end{abstract}

\section{Introduction}

Skin cancer is the most common type, with melanoma being the most aggressive and responsible for 60\% of skin cancer deaths. Early diagnosis is crucial to improve patient survival rates. People of color have a lower risk of developing melanoma than those with lighter skin tones  \cite{acs2022}. 
However, melanin does not entirely protect individuals from developing skin cancer. In fact, acral melanoma, or acrolentiginous melanoma, is the rarest and most aggressive type and occurs more frequently in people with darker skin \cite{AIM}. This subtype is not related to sun exposure, as it tends to develop in areas with low sun exposure, such as the soles, palms, and nails \cite{mskcc2022}. 

When melanoma occurs in individuals with darker skin tones, it is often diagnosed later, making it more challenging to treat and associated with a high mortality rate. This can be partly explained by the fact that acral areas, especially the feet, are often neglected by dermatologists in physical evaluations because they are not exposed to the sun, leading to misdiagnoses \cite{caetano2020melanoma}. Therefore, it is common for melanoma to be confused by patients with fungal infections, injuries, or other benign conditions \cite{mskcc2022}. This is related to the lack of representation of cases of black skin in medical education. Most textbooks do not include images of skin diseases as they appear in black people, or when they do, the number is no more than 10\% \cite{nytimes2020}. This absence can lead to a racial bias in the evaluation of lesions by dermatologists since the same lesion may have different characteristics depending on the patient's skin color\footnote{If you have skin, you can get skin cancer.}, significantly affecting the diagnosis and treatment of these~lesions~\cite{nytimes2020}.

Deep neural networks (DNNs) have revolutionized skin lesion analysis by automatically extracting visual patterns for lesion classification and segmentation tasks. However, training DNNs requires a substantial amount of annotated data, posing challenges in the medical field due to the cost and complexity of data collection and annotation. Transfer learning has emerged as a popular alternative. It involves pre-training a neural network, the encoder, on a large unrelated dataset to establish a powerful pattern extractor. The encoder is fine-tuned using a smaller dataset specific to the target task, enabling it to adapt to skin lesion~analysis.

Despite the advantages of transfer learning, there is a risk that the pre-trained representations may not fully adapt to the target dataset \cite{Menegola2017KnowledgeTF}. 
Self-supervised learning (SSL) has emerged as a promising solution. In SSL, the encoder is trained in a self-supervised manner on unlabeled data using pretext tasks with synthetic labels. The pretext task is only used to stimulate the network to create transformations in the images and learn the best (latent) representations in the feature space that describe them. This way, we have a powerful feature extractor network that can be used in some other target task of interest, i.e., downstream task. Furthermore, applying SSL models for diagnosing skin lesions has proven advantageous, especially in scenarios with scarce training data \cite{chaves2021evaluation}. 

However, deep learning models encounter challenges related to generalization. The effectiveness of machine learning models heavily relies on the quality and quantity of training data available. Unfortunately, in the current medical landscape, skin lesion datasets often suffer from a lack of diversity, predominantly comprising samples from individuals with white skin or lacking explicit labels indicating skin color. This presents a significant challenge as it can lead to models demonstrating racial biases, performing better in diagnosing lesions that are well-represented in the training data from white individuals while potentially encountering difficulties in accurately diagnosing lesions on black skin.

\begin{figure}[t]
\centering
\subfloat[\hspace{1.3cm}(b)\hspace{1.4cm}(c)\hspace{1.4cm}(d)\hspace{1.4cm}(e)\hspace{1.4cm}(f)]{\includegraphics[width=4.25in, clip, trim={0 5cm 0 0}]{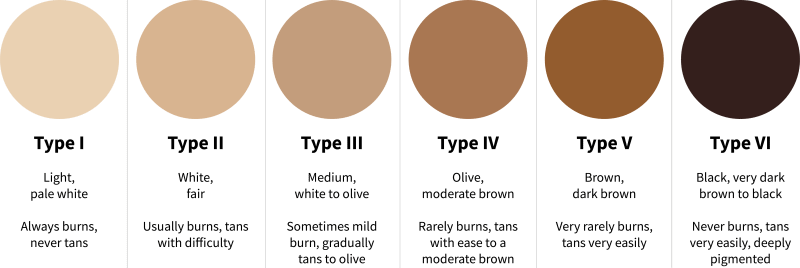}}
\caption{The Fitzpatrick skin type scale. (a) Type 1 (light): pale skin, always burns, and never tans; (b) Type 2 (white): fair, usually burns, tans with difficulty; (c)~Type 3 (medium): white to olive, sometimes mild burn, gradually tans to olive; Type 4 (olive): moderate brown, rarely burns, tans with ease to moderate brown; Type 5 (brown): dark brown, very rarely burns, tans very easily; Type 6 (black): very dark brown to black, never burns, tans very easily, deeply pigmented.}
\label{fig:fitzpatrick-scale}
\end{figure}

\begin{figure}[t]
\centering
\includegraphics[width=0.95\columnwidth]{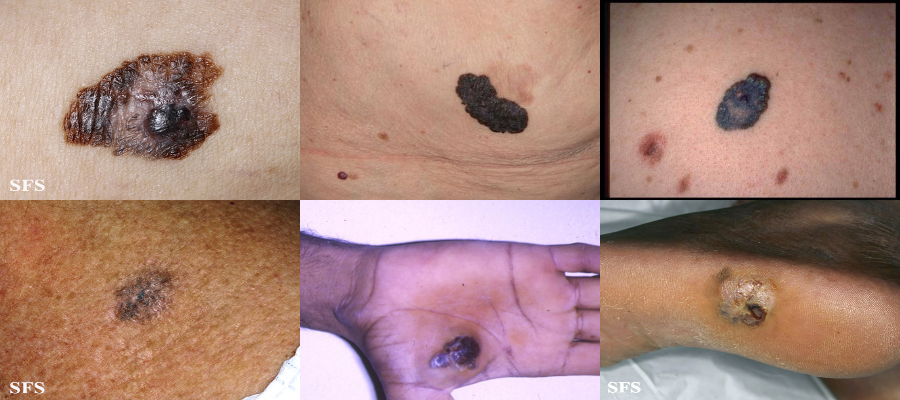}
\caption{Each image corresponds to a melanoma sample and is associated with a specific Fitzpatrick scale value, representing a range of skin tones. The images are organized from left to right, following the Fitzpatrick scale (1 to 6). Images retrieved from Fitzpatrick 17k dataset \cite{groh2021evaluating}.}

\label{fig:lesions}
\end{figure}

Evaluating skin cancer diagnosis models on black skin lesions is one step towards ensuring inclusivity and accuracy across diverse populations \cite{singh2020decolonising}. Most available datasets suffer from insufficient information regarding skin tones, such as the Fitzpatrick scale --- a classification of skin types from 1 to 6 based on a person's ability to tan and their sensitivity and redness when exposed to the sun~\cite{fitzpatrick} (Figs.~\ref{fig:fitzpatrick-scale} and \ref{fig:lesions}). Consequently, we had to explore alternative approaches to address this issue, leading us to conduct the evaluation based on both skin tone and lesion location. We performed two distinct analyses: one focused on directly assessing the impact of skin color using the Fitzpatrick scale, and another centered around evaluating lesions in acral regions, which are more commonly found in individuals with black skin \cite{dln2004}.

The primary objective of this work is to assess the performance of skin cancer classification models, which have performed well in white individuals, specifically on black skin lesions. Our contribution is threefold:
\begin{itemize}
  \item We carefully curate a dataset comprising clinical and dermoscopic images of skin lesions in acral areas (e.g., palms, soles, and nails). 
  \item We evaluate deep neural network models previously trained in a self-super\-vised and supervised manner to diagnose melanoma and benign lesions regarding two types of analysis: 
  \begin{itemize}
       \item Analysis \#1 -- Skin Lesions on Acral Regions: We select images from existing datasets focusing on acral regions.
       \item Analysis \#2 -- Skin Lesions in People of Color: We evaluate datasets that contain Fitzpatrick skin type information.
       
    
     \end{itemize}
  \item We have made the curated sets of data and source code available at \url{https://github.com/httplups/black-acral-skin-lesion-detection}.
\end{itemize}


\section{Related Work}

The accurate diagnosis of skin lesions in people of color, particularly those with dark skin, has been a long-standing challenge in dermatology. One major contributing factor to this issue is the underrepresentation of dark skin images in skin lesion databases. Consequently, conventional diagnostic tools may exhibit reduced accuracy when applied to this specific population, leading to disparities in healthcare outcomes. 

We present a pioneering effort to extensively curate and evaluate the performance of supervised and self-supervised pre-trained models, specifically on black skin lesions and acral regions. While skin lesion classification on acral regions has been explored in previous literature, the focus is largely on general skin types, with limited attention given to black skin tones. Works such as \cite{yu2018acral, lee2020augmented, abbas2021acral} investigated classification performance on acral regions, but they do not specifically address the challenges posed by black skin tones.

Addressing the crucial issue of skin type diversity, Alipour et al.~\cite{alipour2023skin} conducted a comprehensive review of publicly available skin lesion datasets and their metadata. They observed that only PAD-UFES-20~\cite{padufes20}, DDI \cite{doi:10.1126/sciadv.abq6147}, and Fitzpatrick 17k~\cite{groh2021evaluating} datasets provide the Fitzpatrick scale as metadata, highlighting the need for improved representation of diverse skin types in skin lesion datasets. However, the authors did not conduct model evaluations on these datasets.

Existing works explored the application of the Fitzpatrick scale in various areas, such as debiasing~\cite{bevan2022detecting, pakzad2022circle} and image generation~\cite{rezk2022improving}. However, these studies have not adequately addressed the specific challenge of skin lesion classification on black skin tones.

To bridge this research gap, our study evaluates the performance of supervised and self-supervised pre-trained models exclusively on black skin lesion images and acral regions. By systematically exploring and benchmarking different pre-training models, we aim to contribute valuable insights and advancements to the field of dermatology, particularly in the context of underrepresented skin~types.





\section{Materials and Methods}
\label{section:materials-and-methods}

In this work, we assess the performance of six pre-trained models on white skin in black skin. We pre-train all models as described in Chaves et al.~\cite{chaves2021evaluation}. First, we take a pre-trained model backbone on ImageNet~\cite{deng2009imagenet} and fine-tune it on the ISIC dataset~\cite{isic}. The ISIC (\textit{International Skin Imaging Collaboration}) is a common choice in this domain~\cite{Menegola2017KnowledgeTF,valle2020data, groh2021evaluating}, presenting only white skin images. Next, we evaluate the fine-tuned model on several \textbf{out-of-distribution datasets}, where the distribution of the test data diverges from the training one. We also use the same six pre-trained models as Chaves et al.~\cite{chaves2021evaluation} because they have the code and checkpoint publicity available to reproduce their results. The authors compared the performance of five self-supervised models against a supervised baseline and showed that self-supervised pre-training outperformed traditional transfer learning techniques using the ImageNet dataset.
 
We use the ResNet-50 \cite{DBLP:journals/corr/HeZRS15} network as the feature extractor backbone. The self-supervised approaches vary mainly in the choice of pretext tasks, which are BYOL (\textit{Bootstrap Your Own Latent}) \cite{NEURIPS2020_f3ada80d}, InfoMin \cite{NEURIPS2020_4c2e5eaa}, MoCo (\textit{Momentum Contrast}) \cite{9157636}, SimCLR (\textit{Simple Framework for Contrastive Learning of Visual Representations}) \cite{DBLP:journals/corr/abs-2002-05709}, and SwAV (\textit{Swapping Assignments Between Views}) \cite{DBLP:journals/corr/abs-2006-09882}.
We assessed all six models using two different analyses on compound datasets. The first analysis focused on skin lesions in acral regions, while the second considered variations in skin tone. Next, we detail the datasets we curated. 

\subsection{Datasets}

\subsubsection{\textbf{Analysis \#1: Skin Lesions on Acral Regions.}}
To create a compound dataset of acral skin lesions, we extensively searched for datasets and dermatological atlases available on the Internet that provided annotations indicating the location of the lesions. We analyzed 17 datasets listed in SkinIA's website\footnote{\url{https://www.medicalimageanalysis.com/data/skinia}} then filtered the datasets to include only images showcasing lesions in acral regions, such as the palms, soles, and nails. As a result, we identified three widely recognized datasets in the literature, namely the International Skin Imaging Collaboration (ISIC Archive) \cite{isic}, the 7-Point Checklist Dermatology Dataset (Derm7pt) \cite{Kawahara2018-7pt}, and the PAD-UFES-20 dataset \cite{padufes20}. We also included three dermatological atlases:  Dermatology Atlas (DermAtlas) \cite{dermatlas.net}, DermIS \cite{dermis}, and DermNet~\cite{dermnetnz}. 

We describe the steps followed for each dataset in the following. Table~\ref{table:datasets} shows the number of lesions for each dataset.

\begin{description}
    \item[ISIC Archive~\cite{isic}:] We filtered images from the ISIC Archive based on clinical attributes, focusing on lesions on palms and soles, resulting in 773 images. We excluded images classified as carcinoma or unknown, reducing the dataset to 400. As we trained our models using ISIC Archive, we removed all images appearing in the models' training set to avoid data leakage between training and testing data and ensure an unbiased evaluation, resulting in a final dataset with 149 images.\vspace{0.25cm}
    
    \item[Derm7pt \cite{Kawahara2018-7pt}:] It consists of 1011 images for each lesion, including clinical and dermoscopic versions\footnote{Clinical images can be captured with standard cameras, while dermoscopic images are captured with a device called dermatoscope, that normalize the light influence on the lesion, allowing to capture deeper details.}. It offers valuable metadata such as visual patterns, lesion location, patient sex, difficulty level, and 7-point rule scores \cite{argenziano1998epiluminescence}. We applied a filter based on lesion location to select images from it, selecting acral images from the region attribute. This filter resulted in a total of 62 images, comprising only benign and melanoma lesions. We conducted separate evaluations using the clinical and dermoscopic images, labeling the datasets as \textit{derm7pt-clinic} and \textit{derm7pt-derm}, respectively.\vspace{0.25cm} 
    
    \item[PAD-UFES-20 \cite{padufes20}:] It comprises 2298 clinical images collected from smartphone patients. It also includes metadata related to the Fitzpatrick scale, providing additional information about skin tone. We focused on the hand and foot region lesions, which yielded 142 images. We also excluded images classified as carcinoma (malignant), resulting in a final set of 98 images.\vspace{0.25cm} 

   \item[Atlases (DermAtlas, DermIS, DermNet):] The dataset included images obtained from dermatological atlas sources such as DermAtlas \cite{dermatlas.net}, DermIS~\cite{dermis}, and DermNet\cite{dermnetnz}. We use specific search terms, such as \textit{hand}, \textit{hands}, \textit{foot}, \textit{feet}, \textit{acral}, \textit{finger}, \textit{nail}, and \textit{nails} to target the lesion location. We conducted a manual selection to identify images meeting the melanoma or benign lesions criteria. This dataset comprised 10 images from DermAtlas (including 1~melanoma), 13 images from DermIS (comprising 11 melanomas), and 34~images from DermNet, all melanomas. Finally, we combined all images in a set referenced as Atlases, containing 57 images.
\end{description}

\begin{table}[!ht]
\centering
\caption{Number of benign and melanoma lesions for acral areas dataset.}
\label{table:datasets}
\begin{tabular}{lrrr}
\midrule
 & \multicolumn{2}{c}{~~Number of Lesions} & \\

Dataset & Melanoma & Benign & ~~Total\\
\midrule
ISIC Archive \cite{isic} & 72 & 77 & 149 \\
Derm7pt-clinic \cite{Kawahara2018-7pt} & 3 & 59 & 62 \\
Derm7pt-derm \cite{Kawahara2018-7pt} & 3 & 59 & 62 \\
PAD-UFES-20 \cite{padufes20}~~  & 2 & 96 & 98 \\
Atlases \cite{dermatlas.net,dermis,dermnetnz} &  46 & 11  & 57 \\
\midrule
\end{tabular}
\end{table}
\vspace{-.5cm}

\subsubsection{\textbf{Analysis \#2: Skin Lesions in People of Color.}}
We focused on selecting datasets that provided metadata indicating skin tone to analyze skin cancer diagnosis performance for darker-skinned populations. Specifically, datasets containing skin lesions with darker skin tones (Fitzpatrick scales 4, 5, and 6) allow us to evaluate the performance of the models on these populations. For this purpose, we evaluated three datasets: PAD-UFES-20 \cite{padufes20}, which was previously included in the initial analysis, as well as Diverse Dermatology Images (DDI) \cite{doi:10.1126/sciadv.abq6147}, and Fitzpatrick 17k \cite{groh2021evaluating}. 

Table~\ref{table:datasets_fitz} shows the number of lesions for each dataset, considering the Fitzpatrick scale.

\begin{table}[ht]
\renewcommand{\arraystretch}{1.1}
\centering
\caption{Number of benign and melanoma lesions grouped by Fitzpatrick scale for skin tone analysis datasets.}
\label{table:datasets_fitz}
\begin{tabular}{lrrrr} 

\midrule
& Fitzpatrick & \multicolumn{2}{r}{~~~Number of Lesions} & \\
Dataset  & Scale & ~~Melanoma & Benign & ~~Total\\
\midrule

\multirow{4}{*}{PAD-UFES-20* \cite{padufes20}} & 1--2 & 38 & 246 & 284 \\
 & 3--4 & 14 & 153 & 167 \\
 & 5--6 & 0 & 6 & 6 \\ \cmidrule(ll){2-5}
 &Total &52 & 405 & 457  \\ \midrule
 
\multirow{4}{*}{DDI \cite{doi:10.1126/sciadv.abq6147}} & 1--2 & 7 & 153 & 160 \\
 & 3--4 & 7 & 153 & 160 \\
 & 5--6 & 7 & 134 & 141 \\ \cmidrule(ll){2-5}
 &Total &21 & 440 & 461  \\ \midrule

\multirow{4}{*}{Fitzpatrick 17k \cite{groh2021evaluating}} & 1--2 & 331 & 1115 & 1446 \\
 & 3--4 & 168 & 842 & 1010 \\
 & 5--6 & 47 & 203 & 250 \\ \cmidrule(ll){2-5}
 &Total &546 & 2160 & 2706  \\ \midrule

\end{tabular}
\end{table}

\begin{description}
    \item [PAD-UFES-20*:] We filtered images using the Fitzpatrick scale, including lesions from all regions rather than solely acral areas. We specifically selected melanoma cases from the malignant lesions category, excluding basal and squamous cell carcinomas. Also, we excluded images lacking Fitzpatrick scale information. Consequently, the dataset was refined to 457 images, including 52 melanoma cases. Notably, within this dataset, there were only five images with a Fitzpatrick scale of 5 and one image with a Fitzpatrick scale of 6.\vspace{0.25cm}
    
    \item[Diverse Dermatology Images (DDI) \cite{doi:10.1126/sciadv.abq6147}:] The primary objective of DDI is to address the lack of diversity in existing datasets by actively incorporating a wide range of skin tones. For that, the dataset was curated by experienced dermatologists who assessed each patient's skin tone based on the Fitzpatrick scale. The initial dataset comprised 656 clinical images, categorized into different Fitzpatrick scale ranges. We filtered to focus on melanoma samples for malignant lesions. As a result, we excluded benign conditions that do not fall under benign skin lesions, such as inflammatory conditions, scars, and hematomas. This process led to a refined dataset of 461 skin lesions, comprising 440 benign lesions and 21 melanomas. Regarding the distribution based on the Fitzpatrick scale, the dataset includes 160 images from scales~1 to 2, 160 images from scales 3 to 4, and 141 images from scales 5 to 6. The DDI dataset represents a notable improvement in diversity compared to previous datasets, but it still exhibits an unbalanced representation of melanoma images across different skin tones.\vspace{0.25cm}
   
    \item[Fitzpatrick 17k \cite{groh2021evaluating}:] It comprises 16,577 clinical images, including skin diagnostic labels and skin tone information based on the Fitzpatrick scale. The dataset was compiled by sourcing images from two online open-source dermatology atlases: 12,672 images from DermaAmin \cite{dermaamin} and 3,905 images from Atlas Dermatologico \cite{atlasdermatologico}. To ensure the analysis specifically targeted benign and melanoma skin lesion conditions, we applied a filter based on the ``nine\_partition\_attribute''. This filter allowed us to select images that fell into benign dermal, benign epidermal, benign melanocyte, and malignant melanoma. After removing images with the unknown Fitzpatrick value, the refined dataset consists of 2,706 images, 191 images corresponding to a Fitzpatrick scale of 5 and 59 images corresponding to a Fitzpatrick scale of 6. 
\end{description}

\subsection{Evaluation Pipeline}
\begin{figure}[ht]
\centering
\includegraphics[width=\textwidth, trim={0cm 17cm 0 3cm},clip]{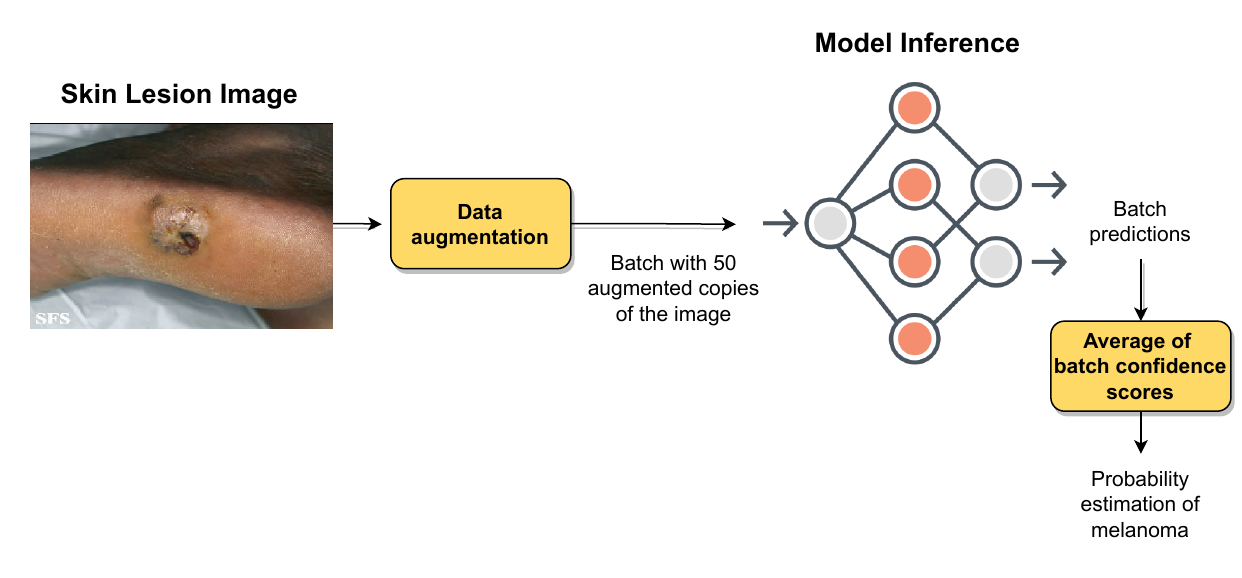}
\caption{Evaluation pipeline for all models. Given a test image, we adopt the final confidence score as the average confidence over a batch of 50 augmented copies of the input image.}
\label{fig:pipeline}
\end{figure}

Our pipeline to evaluate skin lesion image classification models is divided into two main stages: pre-processing and model inference. Fig.~\ref{fig:pipeline} shows the pipeline.

\noindent\textbf{Pre-processing:} We apply data augmentation techniques to the test data, which have been proven to enhance the performance of classification problems~\cite{valle2020data}. The test set is evaluated in batches, and a batch of 50 copies is created for each image. Each copy undergoes various data augmentations, including resizing, flipping, rotations, and color changes. Additionally, we normalize the images using the mean and standard deviation values from the ImageNet dataset.\vspace{0.25cm}

\noindent\textbf{Model inference:} 
The batch of augmented images is fed into the selected model for evaluation. The model generates representations or features specific to its pre-training method. These representations are then passed through a softmax layer, which produces the probability values for the lesion being melanoma, the positive class of interest. We calculate the average of the probabilities obtained from all 50 augmented copies to obtain a single probability value for each image in the batch.\vspace{0.25cm}

The evaluation process consisted of two analyses: Analysis \#1 (skin lesions on acral regions), which considered acral images, and Analysis \#2 (skin lesions in people of color), which considered images with diverse skin tones according to the Fitzpatrick scale. For each analysis, we assessed each dataset individually using the six models: BYOL, InfoMin, MoCo, SimCLR, SwAV, and the Supervised baseline. In each evaluation, the dataset was passed to the respective model, and the probability of melanoma lesions was obtained for all images. Metrics such as balanced accuracy, precision, recall, and F1-score were calculated based on these probabilities. We computed balanced accuracy using a threshold of 0.5.

\section{Results}
\label{section:results}
\subsection{Skin Lesion Analysis on Acral Regions}
Table \ref{table:all_metrics_acral } shows the classification metrics grouped by datasets of SSL models and the supervised baseline for skin lesions in acral regions, such as palms, soles, and nails. In the following, we discussed the results considering each dataset.

\begin{table}[h]
\renewcommand{\arraystretch}{1.0}
\centering
\caption{Evaluation metrics for acral skin lesions. We grouped ISIC Archive, Derm7pt, Atlases, and PAD-UFES-20 due to some datasets' low number of Melanoma samples. \#Mel and \#Ben indicate the number Melanomas, and benign skin lesions, respectively.}
\label{table:all_metrics_acral }
\begin{tabular}{@{\hspace{1.1mm}}clccccc@{\hspace{1.1mm}}} 
\midrule
Samples &  & Balanced\\
(\#Mel/\#Ben) & Model & Accuracy (\%) & ~Precision (\%) & ~Recall (\%)~ & F1-score (\%)\\
\midrule
\multirow{7}{*}{(126/302)} & SwAV & 78.3 & \textbf{77.9} & 64.3 & 70.4 \\
 & MoCo & 79.9 & 72.0 & \textbf{71.4} & 71.7 \\
 & SimCLR & 76.1 & 70.2 & 63.5 & 66.7 \\
 & BYOL & 77.9 & 70.8 & 67.5 & 69.1 \\
 & InfoMin & \textbf{80.2} & 74.2 & 70.6 & \textbf{72.4} \\
 & Supervised & 78.4 & 73.7 & 66.7 & 70.0 \\ \cmidrule{2-6}
 & Mean & 78.5 & 73.1 & 67.3 & 70.1 \\ \midrule

\end{tabular}
\end{table}

\paragraph{ISIC Archive:} We observed consistent result between balanced accuracy and F1-score, both averaging around 87\%. 
The evaluation metrics exhibit high performance due to the fine-tuning process of the evaluated models using the ISIC 2019 dataset. The distribution of the ISIC Archive dataset closely resembles that of the training data, distinguishing it from other datasets, and contributing to the favorable evaluation metrics observed, even though excluding training samples from our evaluation set. Furthermore, in the ISIC~2019 dataset, all results were above 90\% \cite{chaves2021evaluation}. This indicates that even with an external dataset with a distribution more akin to the training data, the performance for lesions in acral regions is significantly inferior to that in other regions. Additionally, it is essential to highlight that in the ISIC 2019 dataset, all results exceeded 90\% \cite{chaves2021evaluation}. 

\paragraph{Derm7pt:} We analyzed two types of images: dermoscopic (derm7pt-derm) and clinical (derm7pt-clinical). When examining the F1-score results for clinical images, the models (SwAV, BYOL, and Supervised) encountered challenges in accurately classifying melanoma lesions. However, the evaluation was performed on a limited sample size of only three melanoma images. This scarcity of data for melanoma evaluation has contributed to the observed zero precision and recall scores. On average, dermoscopic images demonstrated better classification performance than clinical images, with dermoscopic images achieving an F1-score of 26\% and clinical images achieving an F1-score of 16\%. We attribute this disparity to the models being trained on dermoscopic images from the ISIC 2019 dataset. Additionally, using different image capture devices (dermatoscope vs.~cell phone camera) can introduce variations in image quality and the level of detail captured, affecting the overall data distribution. Given that the models were trained with dermoscopic images and the test images were captured using a dermatoscope, the training and test data distributions are expected to be more similar. In general, the results for this dataset demonstrated low F1-score and balanced accuracy, indicating an unsatisfactory performance, especially for clinical images.

\paragraph{Atlases:} The performance varies across different models. MoCo and InfoMin achieved balanced accuracies of approximately 72\%, indicating relatively better performance. Other models, such as Supervised and BYOL, exhibited poor results. Such dataset is considered challenging as it consists of non-standardized skin lesions collected from online atlases, which may introduce variability in the capture process. Still, models could perform better than previous datasets on acral region images, specifically when considering F1-score~values.

\paragraph{PAD-UFES-20:} The models achieved an average balanced accuracy of around 90\%. The model SwAV performed best, with a balanced accuracy of 95.8\% and an F1-score of 33.3\%. All models showed similar patterns: the F1-score and precision were relatively low, while recall was high (100\%). The high recall was mainly due to the correct prediction of the two melanoma samples in the dataset, which inflated the balanced accuracy score. It indicates that relying solely on balanced accuracy can lead to a misleading interpretation of the results. Also, the small number of positive class samples limits the generalizability of the results and reduces confidence in the evaluation.

\subsection{Skin Lesion Analysis in People of Color}

Table~\ref{table:all_metrics_fitz} shows the evaluation results of the SSL models and the Supervised baseline for datasets containing melanoma and benign black skin lesions. 

 \begin{table}[ht]
 \renewcommand{\arraystretch}{1.1}
 \centering
 \caption{Evaluation metrics for skin tone analysis. \#Mel and \#Ben indicate the number Melanomas, and benign skin lesions, respectively.}
 \label{table:all_metrics_fitz}
 \renewcommand{\arraystretch}{1.0} 
 \begin{tabular}{@{\hspace{1.1mm}}clccccc@{\hspace{1.1mm}}} 
 \midrule
 Dataset &  & Balanced\\
 (\#Mel/\#Ben) & Model & Accuracy (\%) & ~Precision (\%) & ~Recall (\%)~ & F1-score (\%)\\
 \midrule

 \multirow{7}{*}{DDI} & SwAV & 52.9 & 7.5 & 14.3 & 9.8 \\
 \multirow{7}{*}{(21/440)} & MoCo & \textbf{55.8} & \textbf{8.5} & \textbf{23.8} & \textbf{12.5} \\
  & SimCLR & 54.2 & 7.8 & 19.0 & 11.1 \\
  & BYOL & 54.3 & 8.0 & 19.0 & 11.3 \\
  & InfoMin & 54.4 & 8.2 & 19.0 & 11.4 \\
  & Supervised & 48.2 & 2.6 & 4.8 & 3.4 \\ \cmidrule{2-6}
  & Mean & 53.3 & 7.1 & 16.7 & 9.9 \\ \midrule

 \multirow{7}{*}{Fitzpatrick 17k} & SwAV & 57.6 & 40.7 & 24.2 & 30.3 \\
 \multirow{7}{*}{(546/2160)} & MoCo & 59.8 & 38.4 & 32.1 & 34.9 \\
  & SimCLR & 59.3 & 40.7 & 29.5 & 34.2 \\
  & BYOL & 59.3 & 38.4 & 32.1 & 34.9 \\
  & InfoMin & 60.1 & 36.5 & \textbf{35.9} & 36.2 \\
  & Supervised & \textbf{63.4} & \textbf{51.2} & 35.3 & \textbf{41.8} \\ \cmidrule{2-6}
  & Mean & 60.0 & 40.4 & 32.4 & 35.6 \\ \midrule

 \multirow{7}{*}{PAD-UFES-20*~} & SwAV & 57.1 & 25.0 & 23.1 & 24.0 \\
 \multirow{7}{*}{(52/405)} & MoCo & 59.1 & 23.9 & 30.8 & 26.9 \\
  & SimCLR & 58.4 & 21.0 & \textbf{32.7} & 25.6 \\
  & BYOL & \textbf{59.2} & \textbf{26.3} & 28.8 & \textbf{27.5} \\
  & InfoMin & 54.3 & 16.9 & 23.1 & 19.5 \\
  & Supervised & 58.5 & 23.8 & 28.8 & 26.1 \\ \cmidrule{2-6}
  & Mean & 57.8 & 22.8 & 27.9 & 24.9 \\
 \midrule
 \end{tabular}
 \end{table}

\paragraph{DDI:} revealed poor results regarding balanced accuracy and F1-score for all models. The supervised baseline model performed the worst, with an F1-score of only 3.4\%, while MoCo achieved a slightly higher F1-score of 12.5\%. Although most of the DDI dataset consisted of benign lesions, the performance of all models was considered insufficient. This underscores the significance of a pre-training process incorporating diverse training data, as it enables the models to learn more robust and generalizable representations across different skin tones and lesion types. In addition, this highlights the importance of self-supervised learning in improving performance and diagnostic accuracy, particularly in the context of diverse skin tones.

\paragraph{Fitzpatrick 17k:} 
In contrast to the DDI dataset, the supervised model achieved the highest performance in balanced accuracy (63.4\%) and F1-score (41.8\%). Both self-supervised and supervised models showed similar results for this dataset.

\paragraph{PAD-UFES-20*:} Both self-supervised and supervised models demonstrated comparable performance. The BYOL method achieved the highest balanced accuracy (59.2\%) and F1-score (27.5\%). It is essential to highlight that this dataset did not include any melanoma lesions corresponding to the Fitzpatrick scale of 5 and 6 (see Table \ref{table:datasets_fitz}).
\section{Conclusion}
\label{section:conclusion}
Our evaluation of self-supervised and supervised models on skin lesions in acral regions reveals a significant deficiency in robustness and bias in deep-learning models for out-of-distribution images, especially in darker skin tones. Both Self-super\-vised and Supervised models achieved poor performance in Melanoma classification task compared to white skin only datasets. These results highlight the generalization gap between models trained on white skin and tested on darker skin tones, inviting further work on improving the generalization capabilities of such models. But, we believe that improvements are not only necessary in model designing, but requires richer data to represent specific population or subgroups.    

The results for melanoma diagnosis in acral regions are insufficient and could cause serious social problems if used clinically. Additionally, more samples are needed to improve the metrics calculation and analysis of results. The generalization power of DNNs-based models heavily depends on training data distribution. Therefore, for DNNs-based models to be robust concerning different visual patterns of lesions, training them with datasets that represent the real clinical scenario, including patients with diverse lesion characteristics and skin tones, is necessary. There is an urgent need for the creation of datasets that guarantee data transparency regarding the source, collection process, and labeling of lesions, as well as the reliability of data descriptions and the ethnic and racial diversity of patients, in order to ensure high confidence in the diagnoses made by the models.

The current state of skin cancer datasets is concerning as it impacts the performance of models and can further reinforce biases in diagnosing skin cancer in people of color. Currently, these models cannot be used in a general sense, as they only perform well on lesions in white skin on common regions affected, and their performance may vary significantly for people with different skin tones. Crafting models that are discriminative for diagnoses, yet discriminate against patients' skin tones, is unacceptable. Deep neural networks have great potential to improve diagnosis, especially for populations with limited access to dermatology. However, including black skin lesions 
is extremely necessary for these populations to access the benefits of inclusive technology. 

\subsubsection{Acknowledgements} L. Chaves is funded by Becas Santander/Unicamp – HUB 2022, Google LARA 2021, in part by the Coordenação de Aperfeiçoamento de Pessoal de Nível Superior – Brasil (CAPES) – Finance Code 001. S.~Avila is funded by CNPq 315231/2020-3, FAEPEX, FAPESP 2013/08293-7, 2020/09838-0, H.IAAC 01245.013778/2020-21, and Google Award for Inclusion Research Program 2022 (``Dark Skin Matters: Fair and Unbiased Skin Lesion Models'').

\bibliographystyle{unsrt}

\bibliography{refs}

\section*{Appendix A}

\setcounter{table}{0}
\renewcommand{\thetable}{A\arabic{table}}

In this section, we stratified the results of Table \ref{table:all_metrics_fitz} by 
Fitzpatrick scale. Tables~\ref{table:ddi_metrics_fitz}, \ref{table:fitzpatrick_metrics_fitz}, and \ref{table:pad_ufes_20_metrics_fitz} shows the results for DDI, Fitzpatrick 17k, and PAD-UFES-20* datasets, respectively. 

\begin{table}[h]
\centering
\caption{Evaluation metrics for DDI dataset. \#Mel and \#Ben indicate the number of Melanomas and benign skin lesions, respectively. $^\dag$None of the seven Melanomas at the 3--4 Fitzpatrick scale were classified correctly.}
\label{table:ddi_metrics_fitz}
\renewcommand{\arraystretch}{1.0} 
\begin{tabular}{@{\hspace{0.7mm}}clccccc@{\hspace{0.7mm}}} 
\midrule
Fitzpatrick Scale &  & Balanced\\
(\#Mel/\#Ben) & Model & Accuracy (\%) & ~Precision (\%) & ~Recall (\%)~ & F1-score (\%)\\
\midrule

\multirow{7}{*}{1--2} & SwAV & \textbf{62.7} & \textbf{28.6} & \textbf{28.6} & \textbf{28.6} \\

\multirow{7}{*}{(7/153)} & MoCo & 54.2 & 10.0 & 14.3 & 11.8 \\
 & SimCLR & 62.0 & 22.2 & \textbf{28.6} & 25.0 \\
 & BYOL & 61.3	& 18.2 & \textbf{28.6} & 22.2 \\
 & InfoMin & 54.9 & 12.5 & 14.3 & 13.3 \\
 & Supervised & 54.5 & 11.1 & 14.3 & 12.5 \\ \cmidrule{2-6}
 & Mean & 58.3 &  17.1 & 21.5 & 18.9 \\ \midrule

\multirow{7}{*}{3--4} & SwAV & \textbf{47.7} & 0.0 & 0.0 & 0.0 \\
\multirow{7}{*}{(7/153)} & MoCo & 45.4 & 0.0 & 0.0 & 0.0 \\
 & SimCLR & 46.4 & 0.0 & 0.0 & 0.0 \\
 & BYOL & 45.1 & 0.0 & 0.0 & 0.0 \\
 & InfoMin & 45.1 & 0.0 & 0.0 & 0.0 \\
 & Supervised & 47.1 & 0.0 & 0.0 & 0.0 \\ \cmidrule{2-6}
 & Mean & 46.1 & 0.0$^\dag$ & 0.0$^\dag$ & 0.0$^\dag$ \\ \midrule

\multirow{7}{*}{5--6} & SwAV & 47.8 & 3.8 & 14.3 & 6.1 \\
\multirow{7}{*}{(7/134)} & MoCo &  57.3 & 24.5 & \textbf{51.1} & 33.1 \\
 & SimCLR & 59.4 & 30.2 & 40.4 & 34.5 \\
 & BYOL & 55.0 & 23.3 & 42.6 & 30.1 \\
 & InfoMin & 59.0 & 26.7 & 48.9 & 34.6 \\
 & Supervised & \textbf{64.2} & \textbf{51.1} & 34.3 & \textbf{41.0} \\ \cmidrule{2-6}
 & Mean & 55.0 & 6.9 & 28.6  & 11.1  \\ 
 \midrule

\end{tabular}
\end{table}

\begin{table}[p]
\centering
\caption{Evaluation metrics for Fitzpatrick 17k dataset. \#Mel and \#Ben indicate the number of Melanomas and benign skin lesions, respectively.}
\label{table:fitzpatrick_metrics_fitz}
\renewcommand{\arraystretch}{1.0} 
\begin{tabular}{@{\hspace{1.1mm}}clccccc@{\hspace{1.1mm}}} 
\midrule
Fitzpatrick Scale &  & Balanced\\
(\#Mel/\#Ben) & Model & Accuracy (\%) & ~Precision (\%) & ~Recall (\%)~ & F1-score (\%)\\
\midrule

\multirow{7}{*}{1--2} & SwAV & 59.1 & 54.4 & 24.2 & 33.5 \\
\multirow{7}{*}{(331/1115)} & MoCo & 60.7 & 43.1 & \textbf{35.0} & 38.7 \\
 & SimCLR & 59.8 & 49.0 & 28.4 & 35.9 \\
 & BYOL & 61.6	& 50.2 & 32.9 & 39.8 \\
 & InfoMin & 61.0 & 44.6 & 34.7 & 39.0 \\
 & Supervised & \textbf{63.7} & \textbf{59.8} & 34.1 & \textbf{43.5} \\ \cmidrule{2-6}
 & Mean &  61.0 & 50.2 & 31.6 & 38.4 \\ \midrule

\multirow{7}{*}{3--4} & SwAV & 55.0 & 28.9 & 19.6 & 23.4 \\
\multirow{7}{*}{(168/842)} & MoCo & 59.8 & 29.2 & \textbf{38.1} & 33.1 \\
 & SimCLR & 58.8 & 34.0 & 28.6 & 31.1 \\
 & BYOL & 57.3 & 30.1 & 27.4 & 28.7 \\
 & InfoMin & 59.2 & 30.1 & 34.5 & 32.1 \\
 & Supervised & \textbf{63.0} & \textbf{47.5} & 33.3 & \textbf{39.2} \\ \cmidrule{2-6}
& Mean & 58.8 & 33.3 & 30.2 & 31.3 \\ \midrule

\multirow{7}{*}{5--6} & SwAV & 59.4 & 30.2 & 40.4 & 34.5 \\
\multirow{7}{*}{(47/203)} & MoCo & 57.3 & 24.5 & \textbf{51.1} &  33.1 \\
 & SimCLR & 59.4 & 30.2 & 40.4 & 34.5 \\
 & BYOL & 55.0 & 23.3 & 42.6 & 30.1 \\
 & InfoMin & 59.0 & 26.7 & 48.9 & 34.6 \\
 & Supervised & \textbf{64.2} & \textbf{34.3} & \textbf{51.1} & \textbf{41.0} \\ \cmidrule{2-6}
 & Mean & 59.0 & 28.2 & 45.8 & 34.6 \\ 
 \midrule

\end{tabular}
\end{table}

\begin{table}[p]
\centering
\caption{Evaluation metrics for PAD-UFES-20* dataset. \#Mel and \#Ben indicate the number of Melanomas and benign skin lesions, respectively. $^\ddag$There is no Melanoma at the 5--6 Fitzpatrick scale.}
\label{table:pad_ufes_20_metrics_fitz}
\renewcommand{\arraystretch}{1.0} 
\begin{tabular}{@{\hspace{1.1mm}}clccccc@{\hspace{1.1mm}}} 
\midrule
Fitzpatrick Scale &  & Balanced\\
(\#Mel/\#Ben) & Model & Accuracy (\%) & ~Precision (\%) & ~Recall (\%)~ & F1-score (\%)\\
\midrule

\multirow{7}{*}{1--2} & SwAV & 57.5 & 26.3 & 26.3 & 26.3 \\
\multirow{7}{*}{(38/246)} & MoCo &  61.3 & \textbf{28.6} & \textbf{36.8} & \textbf{32.2} \\
 & SimCLR & \textbf{60.1} & 25.5 & \textbf{36.8} & 30.1 \\
 & BYOL & 59.7 & \textbf{28.6} & 31.6 & 30.0 \\
 & InfoMin & 54.0 & 18.2 & 26.3 & 21.5 \\
 & Supervised & 56.3 & 21.6 & 28.9 & 24.7 \\ \cmidrule{2-6}
 & Mean &  58.2 & 24.8 & 31.1 & 27.5 \\ \midrule

\multirow{7}{*}{3--4} & SwAV & 54.5 & 20.0 & 14.3 & 16.7 \\
\multirow{7}{*}{(14/153)} & MoCo & 51.9 & 11.1 & 14.3 &  12.5 \\
 & SimCLR & 53.2 & 11.5 & 21.4 & 15.0 \\
 & BYOL & 56.8 & 20.0 & 21.4 & 20.7 \\
 & InfoMin & 52.9 & 13.3 & 14.3 & 13.8 \\
 & Supervised & \textbf{61.7} & \textbf{33.3} & \textbf{28.6} & \textbf{30.8} \\ \cmidrule{2-6}
 & Mean & 55.2 & 18.2 & 19.0 & 18.2 \\ \midrule
 \multirow{7}{*}{5--6} & SwAV & 100 & 0.0 & 0.0 & 0.0 \\
 \multirow{7}{*}{(0/6)} & MoCo & 100 & 0.0 & 0.0 & 0.0 \\
  & SimCLR & 100 & 0.0 & 0.0 & 0.0 \\
  & BYOL & 100	& 0.0 & 0.0 & 0.0 \\
  & InfoMin & 83.3 & 0.0 & 0.0 & 0.0 \\
  & Supervised & 100 & 0.0 & 0.0 & 0.0 \\ \cmidrule{2-6}
  & Mean & 97.2 & 0.0$^\ddag$ & 0.0$^\ddag$ & 0.0$^\ddag$ \\ 
  \midrule

\end{tabular}
\end{table}

\end{document}